\documentclass[twoside,11pt]{article}
\usepackage{pgm}

\usepackage[utf8]{inputenc}
\inputencoding{utf8}

\usepackage{hyperref,color,soul,booktabs}
\setulcolor{blue}
\newcommand{\BibTeX}{\textsc{B\kern-0.1emi\kern-0.017emb}\kern-0.15em\TeX}
\usepackage{bm}
\usepackage{verbatim}
\usepackage{pgf, tikz}
\usetikzlibrary{arrows,automata,fit}

\usepgflibrary{shapes.geometric}
\newcommand{\xx}{1}
\newcommand{\yy}{1}

\newcommand{\stages}[2]{\tikz{\node[shape=circle,draw,inner sep=1pt,fill=#1,minimum size=0.5cm]{\scriptsize{$v_{#2}$}};}} 

\newcommand{\leaf}{\tikz{\node[shape=circle,draw,inner sep=1.5pt,fill=white] {};}}
\newcommand\independent{\protect\mathpalette{\protect\independenT}{\perp}}
\def\independenT#1#2{\mathrel{\rlap{$#1#2$}\mkern2mu{#1#2}}}
 
\usepackage{nicefrac}       % compact symbols for 1/2, etc.
\usepackage{microtype}      % microtypography
\usepackage{xcolor}         % colors

\ShortHeadings{Highly Efficient Structural Learning of Sparse Staged Trees}{Leonelli and Varando}

\begin{document}

% Title and authors
\title{Highly Efficient Structural Learning of Sparse Staged Trees}
\author{\Name{Manuele Leonelli} \Email{manuele.leonelli@ie.edu}\\
   \addr School of Science and Technology, IE University, Madrid, Spain \and
   \Name{Gherardo Varando} \Email{gherardo.varando@uv.es}\\
   \addr Image Processing Laboratory, Universitat de Val\`{e}ncia, Val\`{e}ncia, Spain}

\maketitle

% Abstract and keywords
\begin{abstract}%   <- trailing '%' for backward compatibility of .sty file
Several structural learning algorithms for staged tree models, an asymmetric extension of Bayesian networks, have been defined. However, they do not scale efficiently as the number of variables considered increases. Here we introduce the first scalable structural learning algorithm for staged trees, which searches over a space of models where only a small number of dependencies can be imposed. A simulation study as well as a real-world application illustrate our routines and the practical use of such data-learned staged trees.
\end{abstract}
\begin{keywords}
Asymmetric conditional independence; Bayesian networks; Probabilistic graphical models; Staged trees; Structural learning.
\end{keywords}

\section{Introduction}

Probabilistic graphical models, and in particular Bayesian networks (BNs), are nowadays widely used in machine learning to conveniently represent the relationships existing between the components of a random vector. The directed acyclic graph (DAG) associated to a BN represents graphically (symmetric) conditional independence statements, which can be assessed using the d-separation criterion \citep{pearl1988probabilistic}. Although the underlying DAG can be expert-elicited, this is often learned from data using algorithms that explore the space of all possible DAGs \citep[see e.g.][]{scutari2019learns}.

For quite some time it has been noticed that the strict assumption of symmetric conditional independence may be too restrictive to fully represent the relationship between variables in a dataset \citep{boutilier2013context,Chickering1997,Friedman1996}. However, the development and use in practice of probabilistic graphical models embedding \textit{asymmetric} conditional independence has been limited \citep[see][for some recent proposals]{Hyttinen2018,nicolussi2021context,Talvitie2019}. Possible reasons behind the limited use of such models could be: (i) the lack of widely available software; (ii) the complexity of the learning routines; and (iii) the less intuitive visualization of the associated independences which are not explicitly represented by a single graph.

Staged trees \citep{Collazo2018,Smith2008} are probabilistic graphical models which, starting from an event tree, represent any type of asymmetric conditional independence by a partitioning/coloring of the vertices of the tree.  All the model information and the conditional independences can be read directly from the tree and, in particular, from the coloring of the vertices. The \texttt{stagedtrees} R package \citep{Carli2020} provides a user-friendly implementation of a wide array of structural learning and inferential routines to fit staged trees to data. Therefore, two of the main limitations to the use of asymmetric probabilistic graphical models do not apply to staged trees. Furthermore, a wide toolkit of procedures to work with staged trees have been developed, including handling missing data, sensitivity analysis and exploration of  equivalence classes, among others \citep[see][for details]{Collazo2018}.

On the other hand, learning staged trees from data is complex. Although efficient structural learning algorithms for such models have been implemented \citep[e.g.][]{Freeman2011,leonelli2021context,Silander2013}, they can only work with a limited number of variables. The main reason behind this is the explosion of the size of the model search space as the number of variables considered increases. As an illustration, the number of DAGs over 6 binary variables is 3781503, whilst there are $1.20019 \times 10^{44}$ staged trees under the same conditions \citep{Duarte2021}.  Furthermore, the number of DAGs remains constant if variables have more than two levels, whilst the number of staged trees would further increase dramatically.

Recent proposals for efficient structural learning of staged trees look at sub-classes of staged trees, with the aim of reducing the size of the model space. \citet{Carli2020new} defined \textit{naive staged trees} which have the same complexity of a naive BN over the same variables. \citet{leonelli2022structural} considered \textit{simple staged trees} which have a constrained type of partitioning of the vertices. \citet{Duarte2021} defined \textit{CStrees} which only embed symmetric and context-specific types of independence, and not others \citep{Pensar2016}.

One of the first solutions to make structural learning of BNs scalable was to limit the number of parents each variable can have \citep{friedman1999learning,tsamardinos2006max}. This was  imposed not only to restrict the model space of possible DAGs, but it also made sense from an applied point of view since most often only a limited number of variables can be expected to have a direct influence on another. The option of setting a maximum number of parents is also available in the standard \texttt{bnlearn} software \citep{Scutari2010}. 

Here, we define a sub-class of staged trees embedding the same idea of limiting the number of variables that can have a direct influence to another. As we formalize below, this means that the BN representation associated to such a staged tree is sparse, meaning that it has a small number of edges. A structural learning algorithm for this class of staged trees is introduced and its features are explored in an extensive simulation study.

\section{Bayesian Networks and Staged Trees}

Before introducing staged trees, we give a formal definition of BNs. We then describe their relationships with staged tree models.

\subsection{Bayesian Networks}
 Let $G=([p],E)$ be a DAG with vertex set $[p]=\{1,\dots,p\}$ and edge set $E$. Let $\bm{X}=(X_i)_{i\in[p]}$ be categorical random variables with joint mass function $P$ and sample space $\mathbb{X}=\times_{i\in[p]}\mathbb{X}_i$. For $A\subset [p]$, we let $\bm{X}_A=(X_i)_{i\in A}$ and $\bm{x}_A=(x_i)_{i\in A}$ where $\bm{x}_A\in\mathbb{X}_A=\times_{i\in A}\mathbb{X}_i$. We say that $P$ is Markov to $G$ if, for $\bm{x}\in\mathbb{X}$, 
\[
P(\bm{x})=\prod_{k\in[p]}P(x_k \mid \bm{x}_{\Pi_k}),
\]
where $\Pi_k$ is the parent set of $k$ in $G$ and $P(x_k \vert \bm{x}_{\Pi_k})$ is a shorthand for $P(X_k=x_k \vert \bm{X}_{\Pi_k} = \bm{x}_{\Pi_k})$.  It is customary to label the vertices of a BN so to respect the topological order of $G$ and we henceforth assume that $1,2,\dots,p$ is a topological order of $G$.

The ordered Markov condition implies conditional independences of the form
\[
X_i \independent \bm{X}_{[i-1]}\,\vert\, \bm{X}_{\Pi_i}.
\]
Henceforth, $P$ is assumed to be strictly positive.
Let $G$ be a DAG and $P$ Markov to $G$. The \emph{Bayesian network} model (associated to $G$) is 
\[
\mathcal{M}_G = \{P\in\Delta^{\circ}_{\vert\mathbb{X}\vert-1}\,\vert\, P \mbox{ is Markov to } G\}.
\]
where $\Delta^{\circ}_{\vert\mathbb{X}\vert-1}$ is the ($\vert\mathbb{X}\vert-1$)-dimensional open probability simplex.

%When it is of interest a specific linear ordering $\pi$, we denote the graph $G$ with ordering $\pi$ as $G^{\pi}$.

\subsection{Staged Trees}
Differently to BNs, whose graphical representation is a DAG, staged trees visualize conditional independence by means of a colored tree. Let $(V,E)$ be a directed, finite, rooted tree with vertex set $V$, root node $v_0$ and edge set $E$. 
For each $v\in V$, 
let $E(v)=\{(v,w)\in E\}$ be the set of edges emanating
from $v$ and $\mathcal{C}$ be a set of labels. 
%Given a vector $\bm{x}$, we denote with $\bm{x}_{-i}$ the vector $\bm{x}$ without its $i$-th entry.

An $\bf X$-compatible staged tree %to a categorical random vector $\bm{X}$
is a triple $T = (V,E,\theta)$, where $(V,E)$ is a rooted directed tree and:
\begin{enumerate}
    \item $V = {v_0} \cup \bigcup_{i \in [p]} \mathbb{X}_{[i]}$;
		\item For all $v,w\in V$,
$(v,w)\in E$ if and only if $w=\bm{x}_{[i]}\in\mathbb{X}_{[i]}$ and 
			$v = \bm{x}_{[i-1]}$, or $v=v_0$ and $w=x_1$ for some
$x_1\in\mathbb{X}_1$;
\item $\theta:E\rightarrow \mathcal{L}=\mathcal{C}\times \cup_{i\in[p]}\mathbb{X}_i$ is a labelling of the edges such that $\theta(v,\bm{x}_{[i]}) = (\kappa(v), x_i)$ for some 
			function $\kappa: V \to \mathcal{C}$. The function 
			$k$ is called the colouring of the staged tree $T$.
%\item $\kappa: V \to \mathcal{C}$ is a colouring of the vertices.
\end{enumerate}
	If $\theta(E(v)) = \theta(E(w))$ then $v$ and $w$ are said to be in the same 	\emph{stage}. Therefore, the equivalence classes induced by  $\theta(E(v))$
form a partition of the internal vertices of the tree  in \emph{stages}.

Points 1 and 2 above construct a rooted tree where each root-to-leaf path, or equivalently each leaf, is associated to an element of the sample space $\mathbb{X}$.  Then a labeling of the edges of such a tree is defined where labels are pairs with one element from a set $\mathcal{C}$ and the other from the sample space $\mathbb{X}_i$ of the corresponding variable $X_i$ in the tree. By construction, $\bf X$-compatible staged trees are such that two vertices can be in the same stage if and only if they correspond to the same sample space. 

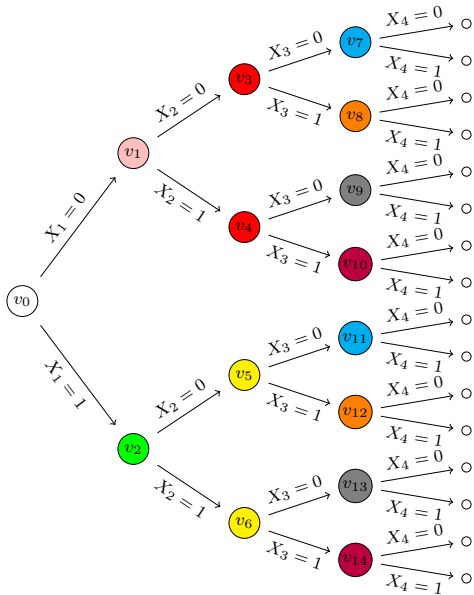
\begin{figure}
\begin{center}
\scalebox{0.82}{
    \begin{tikzpicture}
\renewcommand{\xx}{1.8}
\renewcommand{\yy}{1.2}
\node (v1) at (0*\xx,0*\yy) {\stages{white}{0}};
\node (v2) at (1*\xx,2*\yy) {\stages{pink}{1}};
\node (v3) at (1*\xx,-2*\yy) {\stages{green}{2}};
\node (v4) at (2*\xx,3*\yy) {\stages{red}{3}};
\node (v5) at (2*\xx,1*\yy) {\stages{red}{4}};
\node (v6) at (2*\xx,-1*\yy) {\stages{yellow}{5}};
\node (v7) at (2*\xx,-3*\yy) {\stages{yellow}{6}};
\node (l1) at (3*\xx,3.5*\yy) {\stages{cyan}{7}};
\node (l2) at (3*\xx,2.5*\yy) {\stages{orange}{8}};
\node (l3) at (3*\xx,1.5*\yy) {\stages{gray}{9}};
\node (l4) at (3*\xx,0.5*\yy) {\stages{purple}{10}};
\node (l5) at (3*\xx,-0.5*\yy) {\stages{cyan}{11}};
\node (l6) at (3*\xx,-1.5*\yy) {\stages{orange}{12}};
\node (l7) at (3*\xx,-2.5*\yy) {\stages{gray}{13}};
\node (l8) at (3*\xx,-3.5*\yy) {\stages{purple}{14}};
\node (l9) at (4*\xx,3.75*\yy){\leaf};
\node (l10) at (4*\xx,3.25*\yy){\leaf};
\node (l11) at (4*\xx,2.75*\yy){\leaf};
\node (l12) at (4*\xx,2.25*\yy){\leaf};
\node (l13) at (4*\xx,1.75*\yy){\leaf};
\node (l14) at (4*\xx,1.25*\yy){\leaf};
\node (l15) at (4*\xx,0.75*\yy){\leaf};
\node (l16) at (4*\xx,0.25*\yy){\leaf};
\node (l17) at (4*\xx,-0.25*\yy){\leaf};
\node (l18) at (4*\xx,-0.75*\yy){\leaf};
\node (l19) at (4*\xx,-1.25*\yy){\leaf};
\node (l20) at (4*\xx,-1.75*\yy){\leaf};
\node (l21) at (4*\xx,-2.25*\yy){\leaf};
\node (l22) at (4*\xx,-2.75*\yy){\leaf};
\node (l23) at (4*\xx,-3.25*\yy){\leaf};
\node (l24) at (4*\xx,-3.75*\yy){\leaf};
\draw[->] (v1) --  node [above, sloped] {\scriptsize{$X_1=0$}} (v2);
\draw[->] (v1) -- node [below, sloped] {\scriptsize{$X_1=1$}}(v3);
\draw[->] (v2) --  node [above, sloped] {\scriptsize{$X_2=0$}}(v4);
\draw[->] (v2) --  node [below, sloped] {\scriptsize{$X_2=1$}}(v5);
\draw[->] (v3) --  node [above, sloped] {\scriptsize{$X_2=0$}} (v6);
\draw[->] (v3) --  node [below, sloped] {\scriptsize{$X_2=1$}} (v7);
\draw[->] (v4) --  node [above, sloped] {\scriptsize{$X_3=0$}} (l1);
\draw[->] (v4) -- node [below, sloped] {\scriptsize{$X_3=1$}}  (l2);
\draw[->] (v5) -- node [above, sloped] {\scriptsize{$X_3=0$}}  (l3);
\draw[->] (v5) -- node [below, sloped] {\scriptsize{$X_3=1$}}  (l4);
\draw[->] (v6) -- node [above, sloped] {\scriptsize{$X_3=0$}} (l5);
\draw[->] (v6) -- node [below, sloped] {\scriptsize{$X_3=1$}} (l6);
\draw[->] (v7) -- node [above, sloped] {\scriptsize{$X_3=0$}} (l7);
\draw[->] (v7) -- node [below, sloped] {\scriptsize{$X_3=1$}} (l8);
\draw[->] (l1) --  node [above, sloped] {\scriptsize{$X_4=0$}}(l9);
\draw[->] (l1) --  node [below, sloped] {\scriptsize{$X_4=1$}} (l10);
\draw[->] (l2) --  node [above, sloped] {\scriptsize{$X_4=0$}}(l11);
\draw[->] (l2) --  node [below, sloped] {\scriptsize{$X_4=1$}} (l12);
\draw[->] (l3) --  node [above, sloped] {\scriptsize{$X_4=0$}}(l13);
\draw[->] (l3) --  node [below, sloped] {\scriptsize{$X_4=1$}} (l14);
\draw[->] (l4) --  node [above, sloped] {\scriptsize{$X_4=0$}}(l15);
\draw[->] (l4) --  node [below, sloped] {\scriptsize{$X_4=1$}} (l16);
\draw[->] (l5) -- node [above, sloped] {\scriptsize{$X_4=0$}} (l17);
\draw[->] (l5) --  node [below, sloped] {\scriptsize{$X_4=1$}} (l18);
\draw[->] (l6) -- node [above, sloped] {\scriptsize{$X_4=0$}} (l19);
\draw[->] (l6) --  node [below, sloped] {\scriptsize{$X_4=1$}} (l20);
\draw[->] (l7) -- node [above, sloped] {\scriptsize{$X_4=0$}} (l21);
\draw[->] (l7) --  node [below, sloped] {\scriptsize{$X_4=1$}} (l22);
\draw[->] (l8) -- node [above, sloped] {\scriptsize{$X_4=0$}} (l23);
\draw[->] (l8) -- node [below, sloped] {\scriptsize{$X_4=1$}} (l24);
\end{tikzpicture}
}
\end{center}
\caption{Example of an $\bm{X}$-compatible staged tree over four binary random variables.\label{fig:staged1}}
\end{figure}
Figure \ref{fig:staged1} reports an $(X_1,X_2,X_3,X_4)$-compatible  staged tree over four binary variables. The \textit{coloring} given by the function $\kappa$ is shown in the vertices and
each edge $(\cdot , (x_1, \ldots, x_{i}))$ is labeled with $X_i = x_{i}$. 
The edge labeling $\theta$ can be read from the graph combining the text label and the 
color of the emanating vertex. 
The staging of the staged tree in Figure \ref{fig:staged1} is given by the partition $\{v_0\}$, $\{v_1\}$, $\{v_2\}$, $\{v_3,v_4\}$, $\{v_5,v_6\}$, $\{v_{7},v_{11}\}$, $\{v_{8},v_{12}\}$, $\{v_{9},v_{13}\}$ and $\{v_{10},v_{14}\}$.

The parameter space associated to an $\bf X$-compatible staged tree $T = (V, E, \theta)$ 
with 
labeling $\theta:E\rightarrow \mathcal{L}$ 
is defined as
\begin{equation}
\label{eq:parameter}
	\Theta_T=\Big\{\bm{y}\in\mathbb{R}^{\vert \theta(E)\vert} \;\vert \; \forall ~ e\in E, y_{\theta(e)}\in (0,1)\textnormal{ and }\sum_{e\in E(v)}y_{\theta(e)}=1\Big\}.
\end{equation}
Equation~(\ref{eq:parameter}) defines a class of probability mass functions 
over the edges emanating from any internal vertex coinciding with conditional distributions  $P(x_i \vert \bm{x}_{[i-1]})$, $\bm{x}\in\mathbb{X}$ and $i\in[p]$. In the staged tree in Figure \ref{fig:staged1} the staging $\{v_3, v_4\}$ implies that the conditional distribution of $X_3$ given $X_1=0$ and $X_2 = 0$, represented by the edges emanating from $v_3$, is equal to the conditional distribution of $X_3$ given $X_1=0$ and $X_2=1$. A similar interpretation holds for the staging $\{v_5,v_6\}$. This in turn implies that  $X_3\independent X_2\vert X_1$, thus illustrating that the staging of a tree is associated to conditional independence statements.

Let $\bm{l}_{T}$ denote the leaves of a staged tree $T$. Given a vertex $v\in V$, there is a unique path in $T$ from the root $v_0$ to $v$, denoted as $\lambda(v)$. 	The \emph{depth} of a vertex $v\in V$ equals the number of edges in $\lambda(v)$. For any path $\lambda$ in $T$, let $E(\lambda)=\{e\in E: e\in \lambda\}$ denote the set of edges in the path $\lambda$.

	The \emph{staged tree model} $\mathcal{M}_{T}$ associated to the $\bf X$-compatible staged 
	tree $(V,E,\theta)$ is the image of the map
\begin{equation}
\label{eq:model}
\begin{array}{llll}
\phi_T & : &\Theta_T &\to \Delta_{\vert\bm{l}_T\vert - 1}^{\circ} \\
 &  & \bm{y} &\mapsto \Big(\prod_{e\in E(\lambda(l))}y_{\theta(e)}\Big)_{l\in \bm{l}_T}
\end{array}
\end{equation}

Therefore, staged trees models are such that atomic probabilities are equal to the product of the edge labels in root-to-leaf paths and coincide with the usual factorization of mass functions via recursive conditioning.

\subsection{Staged Trees and Bayesian Networks}

Although the relationship between BNs and staged trees was already formalized 
by \citet{Smith2008}, a formal procedure to represent a BN as a staged tree has been only recently introduced in \citet{duarte2020algebraic} and \citet{varando2021staged}. 

Assume  $\bm{X}$ is topologically ordered with respect to a DAG
$G$ and consider an $\bf X$-compatible staged tree with vertex set $V$,  
edge set $E$ and labeling $\theta$ defined via the 
coloring $\kappa(\bm{x}_{[i]} ) = \bm{x}_{\Pi_{i}}$ of the vertices. The staged tree $T_G$, with vertex set $V$, edge set $E$ and labeling $\theta$
so constructed, is called \emph{the staged tree model of $G$}. 
Importantly,
$\mathcal{M}_G= \mathcal{M}_{T_G}$, i.e. the two models are exactly the same,
since they entail exactly the same factorization of the joint
probability. Clearly, the staging of $T_G$  represents the
Markov conditions associated to the graph $G$.

\citet{varando2021staged} approached the reverse problem of transforming a staged tree into a BN. Of course, since staged trees represent more general asymmetric conditional independences, given a staged tree $T$ most often there is no BN with DAG $G_T$ such that $\mathcal{M}_T=\mathcal{M}_{G_T}$. However,  \citet{varando2021staged} introduced an algorithm that, given an $\bm{X}$-compatible  staged tree $T$, finds the minimal DAG $G_T$ such that $\mathcal{M}_T\subseteq \mathcal{M}_{G_T}$. Minimal means that such a DAG $G_T$ embeds all symmetric conditional independences that are in $\mathcal{M}_T$ and that there are no DAGs with less edges than $G_T$ embedding the same conditional independences.

\begin{figure}
\begin{center}
    \begin{tikzpicture}[
            > = stealth, % arrow head style
            shorten > = 1pt, % don't touch arrow head to node
            auto,
            node distance = 1.5cm, % distance between nodes
            semithick % line style
        ]

        \tikzstyle{every state}=[
            draw = black,
            thick,
            fill = white,
            minimum size = 8mm
        ]

        \node[state] (1) {$1$};
        \node[state] (2) [right of=1] {$2$};
        \node[state] (3) [above of=2] {$3$};
        \node[state] (4) [right of=2] {$4$};
        
        \path[->] (1) edge   (2);
        \path[->] (1) edge   (3);
        \path[->] (3) edge   (4);
        \path[->] (2) edge (4);
 \end{tikzpicture}
\end{center}
\caption{The DAG $G$ such that $T_G$ is the staged tree in Figure \ref{fig:staged1}. \label{fig:bn1}}
\end{figure}
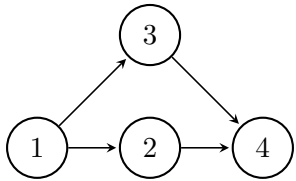

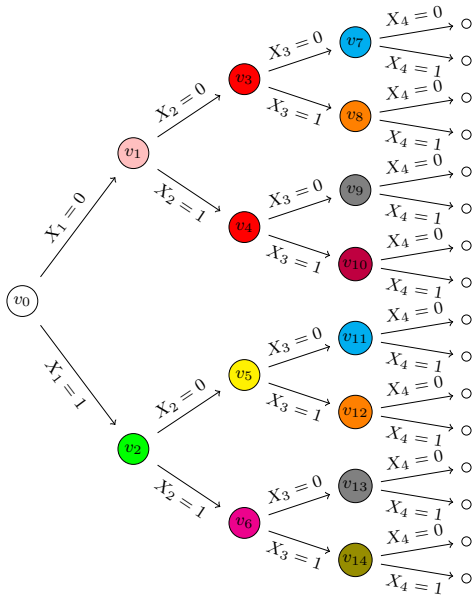
\begin{figure}
\begin{center}
\scalebox{0.82}{
    \begin{tikzpicture}
\renewcommand{\xx}{1.8}
\renewcommand{\yy}{1.2}
\node (v1) at (0*\xx,0*\yy) {\stages{white}{0}};
\node (v2) at (1*\xx,2*\yy) {\stages{pink}{1}};
\node (v3) at (1*\xx,-2*\yy) {\stages{green}{2}};
\node (v4) at (2*\xx,3*\yy) {\stages{red}{3}};
\node (v5) at (2*\xx,1*\yy) {\stages{red}{4}};
\node (v6) at (2*\xx,-1*\yy) {\stages{yellow}{5}};
\node (v7) at (2*\xx,-3*\yy) {\stages{magenta}{6}};
\node (l1) at (3*\xx,3.5*\yy) {\stages{cyan}{7}};
\node (l2) at (3*\xx,2.5*\yy) {\stages{orange}{8}};
\node (l3) at (3*\xx,1.5*\yy) {\stages{gray}{9}};
\node (l4) at (3*\xx,0.5*\yy) {\stages{purple}{10}};
\node (l5) at (3*\xx,-0.5*\yy) {\stages{cyan}{11}};
\node (l6) at (3*\xx,-1.5*\yy) {\stages{orange}{12}};
\node (l7) at (3*\xx,-2.5*\yy) {\stages{gray}{13}};
\node (l8) at (3*\xx,-3.5*\yy) {\stages{olive}{14}};
\node (l9) at (4*\xx,3.75*\yy){\leaf};
\node (l10) at (4*\xx,3.25*\yy){\leaf};
\node (l11) at (4*\xx,2.75*\yy){\leaf};
\node (l12) at (4*\xx,2.25*\yy){\leaf};
\node (l13) at (4*\xx,1.75*\yy){\leaf};
\node (l14) at (4*\xx,1.25*\yy){\leaf};
\node (l15) at (4*\xx,0.75*\yy){\leaf};
\node (l16) at (4*\xx,0.25*\yy){\leaf};
\node (l17) at (4*\xx,-0.25*\yy){\leaf};
\node (l18) at (4*\xx,-0.75*\yy){\leaf};
\node (l19) at (4*\xx,-1.25*\yy){\leaf};
\node (l20) at (4*\xx,-1.75*\yy){\leaf};
\node (l21) at (4*\xx,-2.25*\yy){\leaf};
\node (l22) at (4*\xx,-2.75*\yy){\leaf};
\node (l23) at (4*\xx,-3.25*\yy){\leaf};
\node (l24) at (4*\xx,-3.75*\yy){\leaf};
\draw[->] (v1) --  node [above, sloped] {\scriptsize{$X_1=0$}} (v2);
\draw[->] (v1) -- node [below, sloped] {\scriptsize{$X_1=1$}}(v3);
\draw[->] (v2) --  node [above, sloped] {\scriptsize{$X_2=0$}}(v4);
\draw[->] (v2) --  node [below, sloped] {\scriptsize{$X_2=1$}}(v5);
\draw[->] (v3) --  node [above, sloped] {\scriptsize{$X_2=0$}} (v6);
\draw[->] (v3) --  node [below, sloped] {\scriptsize{$X_2=1$}} (v7);
\draw[->] (v4) --  node [above, sloped] {\scriptsize{$X_3=0$}} (l1);
\draw[->] (v4) -- node [below, sloped] {\scriptsize{$X_3=1$}}  (l2);
\draw[->] (v5) -- node [above, sloped] {\scriptsize{$X_3=0$}}  (l3);
\draw[->] (v5) -- node [below, sloped] {\scriptsize{$X_3=1$}}  (l4);
\draw[->] (v6) -- node [above, sloped] {\scriptsize{$X_3=0$}} (l5);
\draw[->] (v6) -- node [below, sloped] {\scriptsize{$X_3=1$}} (l6);
\draw[->] (v7) -- node [above, sloped] {\scriptsize{$X_3=0$}} (l7);
\draw[->] (v7) -- node [below, sloped] {\scriptsize{$X_3=1$}} (l8);
\draw[->] (l1) --  node [above, sloped] {\scriptsize{$X_4=0$}}(l9);
\draw[->] (l1) --  node [below, sloped] {\scriptsize{$X_4=1$}} (l10);
\draw[->] (l2) --  node [above, sloped] {\scriptsize{$X_4=0$}}(l11);
\draw[->] (l2) --  node [below, sloped] {\scriptsize{$X_4=1$}} (l12);
\draw[->] (l3) --  node [above, sloped] {\scriptsize{$X_4=0$}}(l13);
\draw[->] (l3) --  node [below, sloped] {\scriptsize{$X_4=1$}} (l14);
\draw[->] (l4) --  node [above, sloped] {\scriptsize{$X_4=0$}}(l15);
\draw[->] (l4) --  node [below, sloped] {\scriptsize{$X_4=1$}} (l16);
\draw[->] (l5) -- node [above, sloped] {\scriptsize{$X_4=0$}} (l17);
\draw[->] (l5) --  node [below, sloped] {\scriptsize{$X_4=1$}} (l18);
\draw[->] (l6) -- node [above, sloped] {\scriptsize{$X_4=0$}} (l19);
\draw[->] (l6) --  node [below, sloped] {\scriptsize{$X_4=1$}} (l20);
\draw[->] (l7) -- node [above, sloped] {\scriptsize{$X_4=0$}} (l21);
\draw[->] (l7) --  node [below, sloped] {\scriptsize{$X_4=1$}} (l22);
\draw[->] (l8) -- node [above, sloped] {\scriptsize{$X_4=0$}} (l23);
\draw[->] (l8) -- node [below, sloped] {\scriptsize{$X_4=1$}} (l24);
\end{tikzpicture}
}
\end{center}
\caption{Example of an $\bm{X}$-compatible staged tree over four binary random variables.\label{fig:staged2}}
\end{figure}

As an illustration, the staged tree in Figure \ref{fig:staged1} can be constructed as the $T_G$ from the BN with DAG in Figure \ref{fig:bn1}, embedding the conditional independences $X_3\independent X_2 \;|\;X_1$ and $X_4\independent X_1\;|\;X_2,X_3$. Conversely, consider the staged tree $T$ in Figure \ref{fig:staged2}, which differs from the one in Figure \ref{fig:staged1} only on a different coloring of the vertices $v_6$ and $v_{14}$. Such a staged tree does not embed any symmetric conditional independence, only non-symmetric ones, and therefore there is no DAG $G_T$ such that $\mathcal{M}_{G_T}=\mathcal{M}_T$. Furthermore, the minimal DAG $G_T$ such that $\mathcal{M}_T\subseteq\mathcal{M}_{G_T}$ is the complete one since the staging of the tree implies direct dependence between every pair of variables.

\subsection{Structural Learning Algorithms Inducing Sparsity}
\label{sec:sparse}

Since the model search space of staged trees is huge, we consider here structural learning for a subclass of staged trees that we define next.
\begin{definition}
A staged tree $T$ is in the class of $k$-parents staged trees if the maximum in-degree in $G_T$ is less or equal to $k$.
\end{definition}
For instance, the staged tree in Figure \ref{fig:staged1} is in the class of 2-parents staged trees, whilst the one in Figure \ref{fig:staged2} is not, since its associated minimal DAG is such that $X_4$ has three parents.

Of course, the class of $k$-parents staged trees is much smaller than the one of $\bm{X}$-compatible ones, for small values of $k$, and therefore structural learning is expected to be quicker. Here we define a structural learning algorithm to learn a staged tree in the class of $k$-parents staged trees which consists of the following steps: (i) learn a BN with DAG $G$ having at most $k$ parents (for instance using \texttt{bnlearn}); (ii) construct the equivalent staged tree $T_G$; (iii) run the backward hill-climbing algorithm of \citet{Carli2020} which only joins stages together (no splitting of stages) based on the minimization of the model BIC \citep{gorgen2022curved}. Call the resulting staged tree $T$. It can be easily proven that $G_T$ has at most $k$ parents and $\mathcal{M}_{G_T}\subseteq \mathcal{M}_G$.

Although the idea of using the staged tree equivalent to a BN as starting point of a structural learning algorithm (or at least using a partial ordering associated to such a BN) is not new \citep[see e.g.][]{Barclay2013}, here we specifically use such a strategy to limit the complexity of the learned staged tree. This has two major advantages: (i) the speed of the algorithms increases greatly; (ii) the non-symmetric conditional independences can be easily visualized even when a large number of variables are present, as illustrated in Section \ref{sec:appl}.

\begin{comment}
\textit{The second algorithm does not require the construction of the entire staged tree and consists of the following steps: (i) learn a BN with DAG $G$ having at most $k$ parents; (ii) for each $i=2,\dots,p$, construct the staged tree $T_G$ where $G$ is the DAG induced by $(\bm{X}_{\Pi_i},X_i)$; (iii) for each $i=2,\dots,p$ run the hill-climbing algorithm of \citet{Carli2020} over the vertices associated with $X_i$ only.}

\textit{The second algorithm has the advantage of avoiding constructing the full staged tree $T_G$ which, when the number of variables is large, can be computationally costly and actually may require too much memory. Furthermore, visualize the full staged tree becomes impossible and the staged trees associated with each variable together with the original DAG actually give a graphical representation of every independence of the model. This is similar to the model of \citet{boutilier2013context}, but where the individual trees do not simply represent context-specific independence. On the other hand, the computation of the model BIC is not straightforward and routines to derive it are the topic of ongoing research.}
\end{comment}

\section{Experiments}

We perform simulation experiments to evaluate the proposed 
learning strategy for $k$-parents staged trees.  Moreover we it compare to 
standard learning of staged trees and DAGs.

In all the simulated experiments we generated data from 
random $k$-parents staged trees, which are obtained as follow:
(1) A random DAG $G$ with fixed topological order $X_1, \ldots, X_p$ is obtained
    by randomly selecting up to $k$ parents uniformly from $\{X_1, \ldots, X_{j-1} \}$ for each $X_j$;
(2) The equivalent staged tree $T_G$ is obtained;
(3) Stages in $T_G$ are randomly merged with probability $0.5$. 
The obtained staged tree $T$ is such that $G_T$ is a sub-graph of 
$G$ and thus it is a $k$-parents staged tree;
(4) Lastly, we assign random probabilities (uniformly from the simplex) to each 
stage of the staged tree $T$.
Once we generate the random $k$-parents staged tree $T$, we can easily sample
observations of $X_1, \ldots, X_p$ from it via sequential sampling (as implemented in the \texttt{stagedtrees} package).  
For each fixed parameters ($k$, $p$ and number of observation sampled) we repeat the experiments 
$20$ times and report averages and standard errors.

\subsection{Oracle DAG}

\begin{figure}
    \centering
    \includegraphics{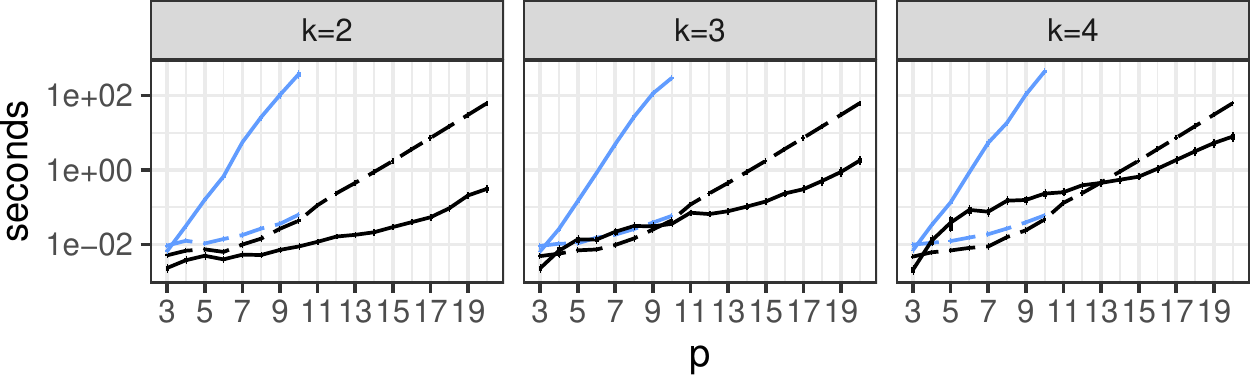}
    \caption{Average computation times for learning the stagedtrees (solid lines) with the standard BHC approach (blue) and the proposed $k$-parents BHC method (black).
    Additionally the time spent creating the initial model are shown (dashed lines). Results for simulated data from $p = 3,\ldots, 20$ binary 
    variables, BHC results are obtained only up to $10$ variables. In different columns results for the maximum number of parents $2,3,4$.}
    \label{fig:time}
\end{figure}

We first consider an ideal scenario where we evaluate the 
performance of the proposed method when the starting DAG is the 
graph $G$ used to generate the true staged tree $T$. 
We thus run a standard backward hill-climbing (BHC) procedure,
as implemented in the \texttt{stagedtrees} package, starting from both the 
full saturated tree model and 
starting from the model $T_G$.
All heuristic searches optimize the BIC score.

We plot in Figure~\ref{fig:time} the average computation time 
as a function of the system size $p=3,\ldots,20$ and varying the number of 
maximum parents $k=2,3,4$. For each one of the two approaches we split the 
computation time in: \texttt{build time}, the time spent building the starting 
tree; and \texttt{search time}, the time spent running the search algorithm. We can observe that starting the search from the 
DAG-equivalent tree allows us to scale the 
algorithm easily up to $20$ variables, while 
the standard approach starting from the full tree 
become quickly infeasible after $10$ variables. 

As a sanity check we also compute the normalized hamming distance (the sum across the depth of the tree of the average number of nodes for which coloring needs to change to obtain the same staged tree) and the 
context intervention distance~\citep{leonelli2021context} between the true and learned models. As expected the $k$-parents trees
obtain better results, which we do not report here for lack of space. 
%Moreover in Figure .... we display the normalized hamming distance between 
%the learned models and the true data-generating staged tree. 

\subsection{Learned DAG}

We perform now a simulation study similar to the oracle setting but, as in a more realistic scenario, we do not assume knowledge of the DAG $G$. 
We thus, first estimate a DAG $\hat{G}$ from data using the hill-climbing approach
in the \texttt{bnlearn} package~\citep{Scutari2010}, and then we apply the BHC
learning algorithm starting from the staged tree $T_{\hat{G}}$ (\texttt{bhcdag}).
We compare the obtained tree to $T_{\hat{G}}$ (the tree equivalent to the learned DAG, \texttt{dag})  and the output of the 
BHC algorithm starting from the full 
saturated model (\texttt{bhc}). 
We run $20$ replications of the experiment for different system sizes ($p=6,10,20$), for different number of maximum 
parents in the true DAG ($k=2,3,4$) and with sample sizes ranging from 
$100$ to $10000$.

\begin{figure}
    \centering
    \includegraphics{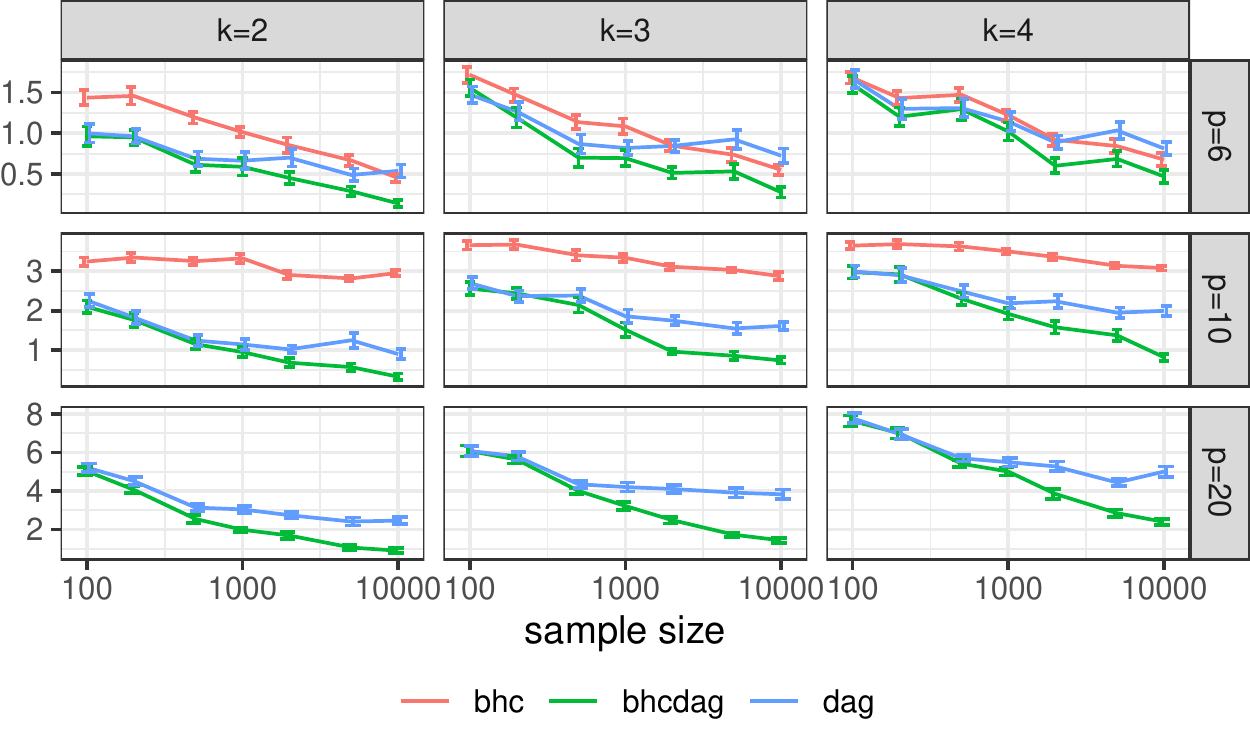}
    \caption{Average normalized hamming distance between true and estimated staged trees as a 
    function of sample size, for different methods, system size ($p$, rows)
    and maximum parents ($k$, columns).}
    \label{fig:hamming}
\end{figure}

In Figure~\ref{fig:hamming} we plot the average (across repetitions) normalized hamming distance 
between the estimated tree and the true one. 
We can observe that the $k$-parents staged trees, obtained by the BHC algorithm
starting from $T_{\hat{G}}$ (\texttt{bhcdag}), are closer to the true 
data-generating models, with respect to both the output of BHC starting from the saturated model and the tree $T_{\hat{G}}$ obtained from the estimated DAG.

\section{COVID-19 Drivers and Country Risks}
\label{sec:appl}
We next extend the analysis of \citet{qazi2022nexus} who developed a BN to investigate how various country risks and risks associated to the COVID-19 epidemics relate to each other. In particular here we focus on how various types of risks affect the overall country risk associated to COVID-19. 

For this purpose, as in \citet{qazi2022nexus}, the dataset used in the analysis comes from the combination of two sources. Country-level exposure to COVID-19 risks are retrieved from INFORM~\citep{inform}. The data comprises of a score between zero and ten for 191 countries for three drivers of COVID-19 risks, namely hazard and exposure, vulnerability, and lack of coping capacity. An overall COVID-19 risk index, again between zero and ten, is constructed from these three drivers. Country-level exposure to various socioeconomic risk factors are collected from Euler Hermes~\citep{euler}. The ratings for five drivers of country risk, namely economic, political, financing, commercial and business environment are collected for 188 countries (the indexes are integer-valued between one and four or six). The combined dataset comprises 181 countries.  Each variable is discretized into two levels using the clustering method from the \texttt{arules} package~\citep{arules}.

\begin{figure}
\begin{center}
    \includegraphics[scale=0.5]{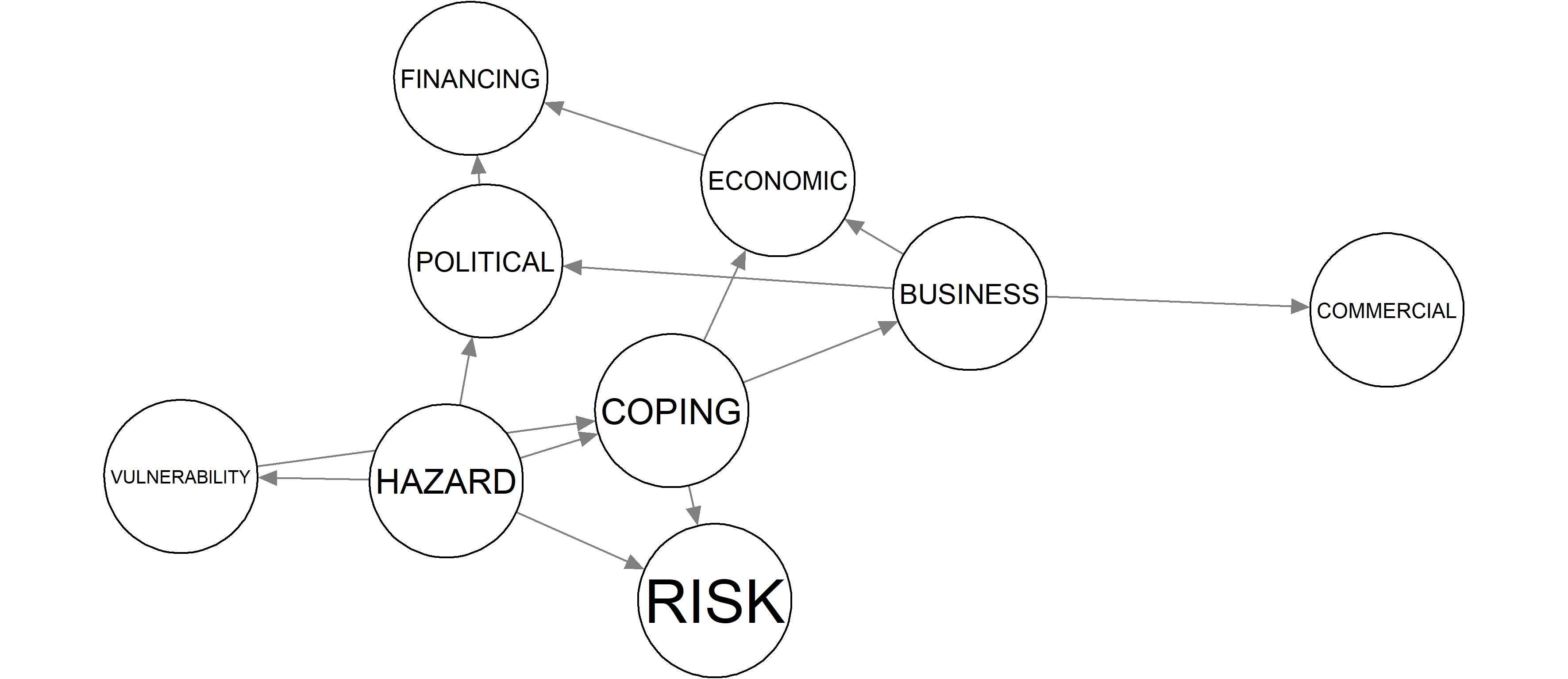}
\end{center}
\caption{BN learned for the COVID-19 drivers and country risks. \label{fig:bncovid}}
\end{figure}

A BN is learned using the \texttt{hc} function of the \texttt{bnlearn} package with the constraint that the overall COVID-19 risk must be a leaf of the network and is reported in Figure \ref{fig:bncovid}. Without specifying it, the learned BN is such that each vertex has at most two parents. The DAG suggests that COVID-19 risk is conditionally independent of all other drivers given the lack of coping capacity and hazard \& exposure. 

\begin{figure}
\begin{center}
\scalebox{0.9}{
    \begin{tikzpicture}
\renewcommand{\xx}{2.8}
\renewcommand{\yy}{0.7}
\node (v1) at (0*\xx,0*\yy) {\stages{white}{0}};
\node (v2) at (1*\xx,2*\yy) {\stages{pink}{1}};
\node (v3) at (1*\xx,-2*\yy) {\stages{green}{2}};
\node (v4) at (2*\xx,3*\yy) {\stages{red}{3}};
\node (v5) at (2*\xx,1*\yy) {\stages{yellow}{4}};
\node (v6) at (2*\xx,-1*\yy) {\stages{cyan}{5}};
\node (v7) at (2*\xx,-3*\yy) {\stages{yellow}{6}};
\node (l1) at (3*\xx,3.5*\yy) {\leaf};
\node (l2) at (3*\xx,2.5*\yy) {\leaf};
\node (l3) at (3*\xx,1.5*\yy) {\leaf};
\node (l4) at (3*\xx,0.5*\yy) {\leaf};
\node (l5) at (3*\xx,-0.5*\yy) {\leaf};
\node (l6) at (3*\xx,-1.5*\yy) {\leaf};
\node (l7) at (3*\xx,-2.5*\yy) {\leaf};
\node (l8) at (3*\xx,-3.5*\yy) {\leaf};
\draw[->] (v1) --  node [above, sloped] {\scriptsize{H = high}} (v2);
\draw[->] (v1) -- node [below, sloped] {\scriptsize{H = low}}(v3);
\draw[->] (v2) --  node [above, sloped] {\scriptsize{C = high}}(v4);
\draw[->] (v2) --  node [below, sloped] {\scriptsize{C = low}}(v5);
\draw[->] (v3) --  node [above, sloped] {\scriptsize{C = high}} (v6);
\draw[->] (v3) --  node [below, sloped] {\scriptsize{C = low}} (v7);
\draw[->] (v4) --  node [above, sloped] {\scriptsize{R = high}} (l1);
\draw[->] (v4) -- node [below, sloped] {\scriptsize{R = low}}  (l2);
\draw[->] (v5) -- node [above, sloped] {\scriptsize{R = high}}  (l3);
\draw[->] (v5) -- node [below, sloped] {\scriptsize{R = low}}  (l4);
\draw[->] (v6) -- node [above, sloped] {\scriptsize{R = high}} (l5);
\draw[->] (v6) -- node [below, sloped] {\scriptsize{R = low}} (l6);
\draw[->] (v7) -- node [above, sloped] {\scriptsize{R = high}} (l7);
\draw[->] (v7) -- node [below, sloped] {\scriptsize{R = low}} (l8);
\end{tikzpicture}
}
\end{center}
\caption{2-parents staged tree for the COVID-19 risk constructed over hazard \& exposure (H), lack of coping capacity (C) and COVID-19 risk (R).\label{fig:staged3}}
\end{figure}
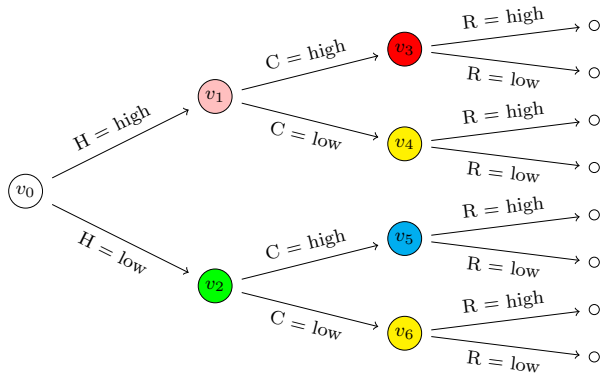

Starting from this BN, a staged tree in the class of 2-parents staged trees is learned using the algorithm of Section \ref{sec:sparse}. This staged tree provides a better representation of the data since it has a BIC of 1521.965, compared to the BIC of 1547.634 of the BN. The learned staged tree embeds the same set of symmetric conditional independences as in the BN of Figure \ref{fig:bn1}, but also non-symmetric ones. Of course, the full tree cannot be easily visualized since, for instance, there are $2^9=512$ vertices with depth nine. However, since it is known that COVID-19 risk only depends on the lack of coping capacity and hazard \& exposure we can construct the staged tree over these three variables only and easily visualize further non-symmetric dependences. This is reported in Figure \ref{fig:staged3}, which shows the presence of the context-specific independence between COVID-19 risk and hazard \& exposure for lack of coping capacity equal to low. Similar interpretations could be drawn by constructing the ``partial" staged trees associated to other variables.

Of course a generic staged tree would provide a better representation of the data. For instance, one learned with the backward hill-climbing of \citet{Carli2020} starting from the saturated model has a BIC of 1453.587. However, its complete visualization is again unfeasible and plots as the one of Figure \ref{fig:staged3} are in general not viable since there are no constraints on the number of parents of $G_T$. Indeed, whilst the DAG $G_T$ for the tree in Figure \ref{fig:staged3} has 13 edges (as the DAG in Figure \ref{fig:bncovid}), the DAG $G_T$ from the generic staged tree is complete and consisting of  36 edges, meaning that all variables are directly related to one another.

\section{Discussion}
We defined a novel sub-class of staged tree models borrowing the idea of limiting the number of parents in BNs. A structural learning algorithm for such a class has been introduced and its properties illustrated both in simulation experiments and in a real-world application.

The number of parents of a variable in the staged tree is limited by those of the learned BN. For instance, in our application, although no limit on the number of parents was set, there were at most two parents. However, non-symmetric dependences may have been missed by the BN learning algorithm which is specifically designed to account for symmetric ones. One possibility could be to add edges to the learned BN, for instance between variables such that their conditional mutual information is large, and then run a backward hill-climbing algorithm of \citet{Carli2020} over that BN. The feasibility of such an algorithm is the focus of current research.

\subsection*{Acknowledgements}

Gherardo Varando's work was funded 
by the European Research 
Council (ERC) Synergy Grant 
“Understanding and Modelling the Earth 
System with Machine Learning (USMILE)” 
under Grant Agreement No 855187.

\bibliography{references}

\begin{thebibliography}{29}
\providecommand{\natexlab}[1]{#1}
\providecommand{\url}[1]{\texttt{#1}}
\expandafter\ifx\csname urlstyle\endcsname\relax
  \providecommand{\doi}[1]{doi: #1}\else
  \providecommand{\doi}{doi: \begingroup \urlstyle{rm}\Url}\fi

\bibitem[Barclay et~al.(2013)Barclay, Hutton, and Smith]{Barclay2013}
L.~M. Barclay, J.~L. Hutton, and J.~Q. Smith.
\newblock {Refining a Bayesian network using a chain event graph}.
\newblock \emph{International Journal of Approximate Reasoning}, 54:\penalty0
  1300--1309, 2013.

\bibitem[Boutilier et~al.(1996)Boutilier, Friedman, Goldszmidt, and
  Koller]{boutilier2013context}
C.~Boutilier, N.~Friedman, M.~Goldszmidt, and D.~Koller.
\newblock {Context-specific independence in Bayesian networks}.
\newblock In \emph{Proceedings of the Twelfth Conference on Uncertainty in
  Artificial Intelligence}, pages 115--123, 1996.

\bibitem[Carli et~al.(2020)Carli, Leonelli, and Varando]{Carli2020new}
F.~Carli, M.~Leonelli, and G.~Varando.
\newblock A new class of generative classifiers based on staged tree models.
\newblock \emph{arXiv:2012.13798}, 2020.

\bibitem[Carli et~al.(2022)Carli, Leonelli, Riccomagno, and Varando]{Carli2020}
F.~Carli, M.~Leonelli, E.~Riccomagno, and G.~Varando.
\newblock The {R} package stagedtrees for structural learning of stratified
  staged trees.
\newblock \emph{Journal of Statistical Software}, 102\penalty0 (6):\penalty0
  1--30, 2022.

\bibitem[Chickering et~al.(1997)Chickering, Heckerman, and
  Meek]{Chickering1997}
D.~M. Chickering, D.~Heckerman, and C.~Meek.
\newblock {A Bayesian approach to learning Bayesian networks with local
  structure}.
\newblock In \emph{Proceedings of 13th Conference on Uncertainty in Artificial
  Intelligence}, pages 80--89, 1997.

\bibitem[Collazo et~al.(2018)Collazo, G\"{o}rgen, and Smith]{Collazo2018}
R.~Collazo, C.~G\"{o}rgen, and J.~Smith.
\newblock \emph{{Chain event graphs}}.
\newblock Chapmann \& Hall, 2018.

\bibitem[Duarte and Solus(2020)]{duarte2020algebraic}
E.~Duarte and L.~Solus.
\newblock Algebraic geometry of discrete interventional models.
\newblock \emph{arXiv:2012.03593}, 2020.

\bibitem[Duarte and Solus(2021)]{Duarte2021}
E.~Duarte and L.~Solus.
\newblock Representation of context-specific causal models with observational
  and interventional data.
\newblock \emph{arXiv:2101.09271}, 2021.

\bibitem[{Euler Hermes}(2022)]{euler}
{Euler Hermes}.
\newblock Country risk reports.
\newblock Retrieved from: https://www.eulerherm
  es.com/en\_global/economic-research/country-reports.html, 2022.

\bibitem[Freeman and Smith(2011)]{Freeman2011}
G.~Freeman and J.~Q. Smith.
\newblock {Bayesian MAP model selection of chain event graphs}.
\newblock \emph{Journal of Multivariate Analysis}, 102\penalty0 (7):\penalty0
  1152--1165, 2011.

\bibitem[Friedman and Goldszmidt(1996)]{Friedman1996}
N.~Friedman and M.~Goldszmidt.
\newblock Learning {B}ayesian networks with local structure.
\newblock In \emph{Proceedings of the 12th Conference on Uncertainty in
  Artificial Intelligence}, pages 252--262, 1996.

\bibitem[Friedman et~al.(1999)Friedman, Nachman, and
  Pe'er]{friedman1999learning}
N.~Friedman, I.~Nachman, and D.~Pe'er.
\newblock {Learning Bayesian network structure from massive datasets: The
  ``sparse candidate" algorithm}.
\newblock In \emph{Proceedings of the 15th Conference on Uncertainty in
  Artificial Intelligence}, pages 206--215, 1999.

\bibitem[G{\"o}rgen et~al.(2022)G{\"o}rgen, Leonelli, and
  Marigliano]{gorgen2022curved}
C.~G{\"o}rgen, M.~Leonelli, and O.~Marigliano.
\newblock The curved exponential family of a staged tree.
\newblock \emph{Electronic Journal of Statistics}, 16\penalty0 (1):\penalty0
  2607--2620, 2022.

\bibitem[Hahsler et~al.(2005)Hahsler, Gruen, and Hornik]{arules}
M.~Hahsler, B.~Gruen, and K.~Hornik.
\newblock arules -- {A} computational environment for mining association rules
  and frequent item sets.
\newblock \emph{Journal of Statistical Software}, 14\penalty0 (15):\penalty0
  1--25, 2005.

\bibitem[Hyttinen et~al.(2018)Hyttinen, Pensar, Kontinen, and
  Corander]{Hyttinen2018}
A.~Hyttinen, J.~Pensar, J.~Kontinen, and J.~Corander.
\newblock {Structure learning for Bayesian networks over labeled DAGs}.
\newblock In \emph{International Conference on Probabilistic Graphical Models},
  pages 133--144, 2018.

\bibitem[{INFORM}(2022)]{inform}
{INFORM}.
\newblock {COVID-19 risk index}.
\newblock Retrieved from: https://drmkc.jrc.
  ec.europa.eu/inform-index/INFORM-Covid-19, 2022.

\bibitem[Leonelli and Varando(2021)]{leonelli2021context}
M.~Leonelli and G.~Varando.
\newblock Context-specific causal discovery for categorical data using staged
  trees.
\newblock \emph{arXiv:2106.04416}, 2021.

\bibitem[Leonelli and Varando(2022)]{leonelli2022structural}
M.~Leonelli and G.~Varando.
\newblock Structural learning of simple staged trees.
\newblock \emph{arXiv:2203.04390}, 2022.

\bibitem[Nicolussi and Cazzaro(2021)]{nicolussi2021context}
F.~Nicolussi and M.~Cazzaro.
\newblock Context-specific independencies in stratified chain regression
  graphical models.
\newblock \emph{Bernoulli}, 27\penalty0 (3):\penalty0 2091--2116, 2021.

\bibitem[Pearl(1988)]{pearl1988probabilistic}
J.~Pearl.
\newblock \emph{Probabilistic reasoning in intelligent systems: networks of
  plausible inference}.
\newblock Morgan Kaufmann, 1988.

\bibitem[Pensar et~al.(2016)Pensar, Nyman, Lintusaari, and
  Corander]{Pensar2016}
J.~Pensar, H.~Nyman, J.~Lintusaari, and J.~Corander.
\newblock The role of local partial independence in learning of {B}ayesian
  networks.
\newblock \emph{International Journal of Approximate Reasoning}, 69:\penalty0
  91--105, 2016.

\bibitem[Qazi and Simsekler(2022)]{qazi2022nexus}
A.~Qazi and M.~C.~E. Simsekler.
\newblock {Nexus between drivers of COVID-19 and country risks}.
\newblock \emph{Socio-Economic Planning Sciences}, page 101276, 2022.

\bibitem[Scutari(2010)]{Scutari2010}
M.~Scutari.
\newblock {Learning Bayesian networks with the bnlearn R package}.
\newblock \emph{Journal of Statistical Software}, 35\penalty0 (3):\penalty0
  1--22, 2010.

\bibitem[Scutari et~al.(2019)Scutari, Graafland, and
  Guti{\'e}rrez]{scutari2019learns}
M.~Scutari, C.~E. Graafland, and J.~M. Guti{\'e}rrez.
\newblock Who learns better {B}ayesian network structures: Accuracy and speed
  of structure learning algorithms.
\newblock \emph{International Journal of Approximate Reasoning}, 115:\penalty0
  235--253, 2019.

\bibitem[Silander and Leong(2013)]{Silander2013}
T.~Silander and T.~Leong.
\newblock A dynamic programming algorithm for learning chain event graphs.
\newblock In \emph{Proceedings of the International Conference on Discovery
  Science}, pages 201--216, 2013.

\bibitem[Smith and Anderson(2008)]{Smith2008}
J.~Smith and P.~Anderson.
\newblock {Conditional independence and chain event graphs}.
\newblock \emph{Artificial Intelligence}, 172\penalty0 (1):\penalty0 42 -- 68,
  2008.

\bibitem[Talvitie et~al.(2019)Talvitie, Eggeling, and Koivisto]{Talvitie2019}
T.~Talvitie, R.~Eggeling, and M.~Koivisto.
\newblock Learning {B}ayesian networks with local structure, mixed variables,
  and exact algorithms.
\newblock \emph{International Journal of Approximate Reasoning}, 115:\penalty0
  69--95, 2019.

\bibitem[Tsamardinos et~al.(2006)Tsamardinos, Brown, and
  Aliferis]{tsamardinos2006max}
I.~Tsamardinos, L.~E. Brown, and C.~F. Aliferis.
\newblock {The max-min hill-climbing Bayesian network structure learning
  algorithm}.
\newblock \emph{Machine Learning}, 65\penalty0 (1):\penalty0 31--78, 2006.

\bibitem[Varando et~al.(2021)Varando, Carli, and Leonelli]{varando2021staged}
G.~Varando, F.~Carli, and M.~Leonelli.
\newblock Staged trees and asymmetry-labeled {DAGs}.
\newblock \emph{arXiv:2108.01994}, 2021.

\end{thebibliography}
\end{document}